\documentclass{article}

\usepackage{arxiv}

\usepackage[utf8]{inputenc} 
\usepackage[T1]{fontenc}    
\usepackage{url}            
\usepackage{booktabs}       
\usepackage{amsfonts}       
\usepackage{nicefrac}       
\usepackage{microtype}      
\usepackage{graphicx}
\graphicspath{ {./images/} }

\newcommand{\etal}{\textit{et al.}}
\newcommand{\ie}{\textit{i.e.}}
\newcommand{\eg}{\textit{e.g.}}

\usepackage{amsmath}
\usepackage{amssymb}
\usepackage{booktabs}
\usepackage{multirow}
\usepackage{caption}
\usepackage{subcaption}
\usepackage{array}

\title{TEN: \textbf{T}win \textbf{E}mbedding \textbf{N}etworks\\ for the Jigsaw Puzzle Problem\\ with Eroded Boundaries}

\author{
    Daniel Rika \\
    Dept.~of Computer Science\\
    Bar-Ilan University\\
    Israel, Ramat-Gan 52900 \\
    \texttt{danielrika@gmail.com}
    \And
    Dror Sholomon \\
    Dept.~of Computer Science\\
    Bar-Ilan University\\
    Israel, Ramat-Gan 52900 \\
    \texttt{dror.sholomon@gmail.com}
    \And
    Eli (Omid) David \\
    Dept.~of Computer Science\\
    Bar-Ilan University\\
    Israel, Ramat-Gan 52900 \\
    \texttt{mail@elidavid.com}
    \And
    Nathan S.~Netanyahu \\
    Dept.~of Computer Science\\
    Bar-Ilan University\\
    Israel, Ramat-Gan 52900 \\
    \texttt{nathan@cs.biu.ac.il}
}

\begin{document}
\maketitle
\begin{abstract}
This paper introduces the novel CNN-based encoder 
\textit{Twin Embedding Network} (TEN), for the jigsaw puzzle problem (JPP), which represents a puzzle piece with respect to its boundary in a latent embedding space. Combining this latent representation with a simple distance measure, we demonstrate improved accuracy levels of our newly proposed pairwise \textit{compatibility measure} (CM), compared to that of various classical methods, for degraded puzzles with eroded tile boundaries. We focus on this problem instance for our case study, as it serves as an appropriate testbed for real-world scenarios.
Specifically, we demonstrated an improvement of up to $8.5\%$ and $16.8\%$ in reconstruction accuracy, for so-called Type-1 and Type-2 problem variants, respectively. Furthermore, we also demonstrated that TEN is faster by a few orders of magnitude, on average, than a typical deep neural network (NN) model, {\ie}, it is as fast as the classical methods. In this regard, the paper makes a significant first attempt at bridging the gap between the relatively low accuracy (of classical methods and the intensive computational complexity (of NN models), for practical, real-world puzzle-like problems.
\end{abstract}

\section{Introduction}

Numerous successful studies have been pursued over the years for solving the intriguing \textit{jigsaw puzzle problem} (JPP). In particular, most advanced methods have focused recently on jigsaw puzzles containing non-overlapping square pieces.
The reconstruction of such puzzles is composed typically of two main steps. First, a \textit{compatibility measure} (CM) is computed for all pairwise piece sides, to obtain numerically the degree of compatibility for each possible piece pair. After computing all of the pairwise CMs, a reconstruction phase kicks in.
The goal of any reconstruction algorithm is to determine correctly the location and orientation of each piece, relying solely on the CM scores computed in the first step.
All previous works vary in the way they compute the pairwise CMs and in the way they try to assemble a puzzle as accurately as possible.
Although the two steps are equally essential for the reconstruction, we focus here on the CM computation of all pairwise pieces, which could be highly expensive due to the order of $\mathcal{O}(N^2)$ operations required. (Note that an $N$-piece puzzle requires $16N^2$ pairwise CM computations.)

Classical, most commonly used CMs include the
\textit{sum of squared distances} (SSD) proposed by Cho~{\etal}~\cite{conf/cvpr/ChoAF10}, the so-called  \textit{prediction-based compatibility} (PBC), employing $(L_p)^q$ variants due to  Pomerantz~{\etal}~\cite{conf/cvpr/PomeranzSB11}, the \textit{$L_1$-norm compatibility} of Paikin and Tal~\cite{paikin2015solving}, and the \textit{Mahalanobis gradient compatibility} (MGC) proposed by Gallagher~\cite{gallagher2012jigsaw}. These and other measures  ({\eg} \cite{son2014solving,yu2016bmvc,son2019tpami}) yield high quality CMs, 
which allow for successful reconstruction of very large puzzles with up to tens of thousands of pieces.

The effectiveness of the above methods greatly depends, however, on the notion of a \textit{synthetic puzzle}, {\ie} a puzzle obtained by cropping a high-resolution image into $N$ non-overlapping square pieces of size $P \times P$ pixels. (Such puzzles contain perfect pieces, and are considered information rich.)  Indeed, the classical methods tend to be simple, fast, and very accurate under this favorable working assumption, 
which bodes well with CMs based mostly on piece boundary pixels.

When encountering, though, harder puzzle problems (due to various \textit{real-world} scenarios, such as monochromatic puzzles, eroded piece boundaries, missing pieces, etc.), the classical measures tend to perform poorly, thereby affecting negatively the reconstruction outcome. Mondal~{\etal}~\cite{mondal2013crv} were the first to  handle this weak spot for puzzles containing \textit{eroded boundaries}. They proposed a new pairwise CM called \textit{weighted-MGC} (wMGC), which combines the SSD and MGC CMs. Despite their slight improvement, their wMGC still suffers from severe degradation in performance for eroded piece boundaries of 1--2 pixels.

The growing influence of \textit{deep neural networks} (DNNs), in many related fields, has given rise recently also to the development of more accurate and robust CMs for harder variants of the JPP.
Pai\~{x}ao~{\etal}~\cite{paixao2018deep} used a pre-trained SqueezeNet~\cite{SqueezeNet} model and fine-tuned it to obtain an effective CM for \textit{strip-cut shredded documents}. Their NN-based pairwise CM enabled them to reassemble mechanically shredded documents with 94\% accuracy. Another real-world JPP variant concerns the reconstruction of \textit{Portuguese tile panels}, where often (square) real panel tiles have been severely degraded chromatically and eroded along their edges over hundreds of years. Rika~{\etal}~\cite{rika2019gecco} proposed an ensemble of four NN-based models for dealing with this challenging variant.
Their NN-based CM provides a significant improvement over classical methods, which made the reconstruction of Portuguese panels feasible for the first time.
Regarding the problem of eroded boundaries, Bridger~{\etal}~\cite{bridger2020cvpr} used a \textit{generative adversarial network} (GAN) to ``in-paint'' the gap between pairs of pieces. After sufficient training of their generator, they keep its weights fixed and apply a second training stage to fine-tune the discriminator to differentiate between a true adjacent in-paint pair and false ones. Inspired by Pix2Pix~\cite{pix2pix2017}, their generator uses a U-Net encoder-decoder architecture~\cite{ronneberger2015unet} and
a Markovian discriminator with a 3-layer encoder. They carried out their work on a piece size $64 \times 64$ pixels, with an erosion of 2 or 4 boundary pixels,
and reported superior results to those of the classical $L_1$-norm compatibility.
Other methods aimed at solving the JPP entirely by DNNs~\cite{deepzzle2020tip,jigsawgan2022tip} were shown to work only on virtually impractical puzzle sizes of $4 \times 4$. 

Unfortunately, DNN-based CMs consist of millions of parameters, such that each inference becomes very computationally-intensive.
Since the model should be employed $16N^2$ times for an $N$-piece puzzle, the entire CM phase becomes highly infeasible in the lack of powerful dedicated hardware.

As shown in Section~\ref{sec:inference_times}, the running time of a highly-intensive NN-based CM grows exponentially, and could take a couple of hours just to compute the pairwise CMs of a puzzle with only a few thousands pieces. Thus, tackling real-world problems like \textit{shredded documents}, \textit{ancient tile panels}, etc., which involve multiple puzzles containing millions of pieces, would render essentially NN-based CMs impractical.

To alleviate significantly this serious deficiency, we propose in this paper a new methodology for computing efficiently a pairwise CM via the \textit{Twin Embedding (Neural) Network} (TEN) framework. The main idea of our novel approach is to employ TEN to represent an \textit{entire puzzle piece} with respect to \textit{each of its boundaries in a latent embedding space}. Combining this representation with a simple distance measure, we then demonstrate that our newly obtained CM is significantly more accurate than well-known classical CMs and, at the same time, is faster by a few orders of magnitude than typical NN-based CMs.   
The main highlights of our work are summarized as follows:
\begin{itemize}
    \item[$\bullet$] Proposed a new NN-based framework and architecture to represent via embeddings an entire puzzle piece with respect to its boundaries;
    \item[$\bullet$] Demonstrated the use of embedding NNs for speeding up the CM computation by a few orders of magnitude, relatively to a standard end-to-end NN-based scheme;
    \item[$\bullet$] Presented extensive Top-1 accuracy (to be defined) and reconstruction results, which demonstrate TEN's superior performance to that of most classical CMs;
    \item[$\bullet$] Conducted extensive empirical tests, involving  220 puzzles (with hundreds of pieces each) of three different datasets, as well as two reconstruction algorithms, to further establish the superior performance of our proposed framework, in comparison to previous CM methods.
\end{itemize}
We believe that the above contributions make a significant first step at bridging the gap between the relative low accuracy (of classical CMs) and the intensive computational complexity (of NN-based models), for various real-world puzzle-like problems.

The rest of the paper is organized as follows. Section 2 describes in detail TEN's framework and architecture. Section 3 presents our extensive experimental results, including a comparative performance evaluation with previous classical CMs. Finally, Section 4 makes concluding remarks and discusses future work. 

\section{TEN: Twin Embedding (Neural) Networks}
\label{sec:twin_embedding_nets}

Traditional methods (not based on NN models) typically compute the distance measure,
$D(e_l, e_r): \mathbb{R}^d \times \mathbb{R}^d \to \mathbb{R}$, for computing the CM of a given pair of tiles, where $e_l$ and $e_r$ correspond, respectively, to the pixels along the left and right tile edges.
Initially, only edge pixels were usually taken into account. More recent methods (like Gallagher's MGC) considered 2-pixel wide edges for more informative processing.
Hence, if a puzzle contains $P \times P \times C$-size pieces, where $P$ and $C$ denote, respectively, the piece dimensions and number of (color) channels, then $e_l, e_r \in \mathbb{R}^{P\times 2 \times C}$.
In contrast, NN-based dissimilarity functions work directly on an \textit{entire tile pair} $[k_l:k_r]$, in an attempt to exploit maximum content information from the pieces $k_l$ and $k_r$, which can be formulated as $g([k_l:k_r]): \mathbb{R}^{P \times 2P \times C} \to \mathbb{R}$.
Thus, the computational complexity associated with these functions is considerably more intensive.

\begin{figure}
  \centering
  \begin{subfigure}{0.3\linewidth}
  \centering
    \includegraphics[width=0.7\linewidth]{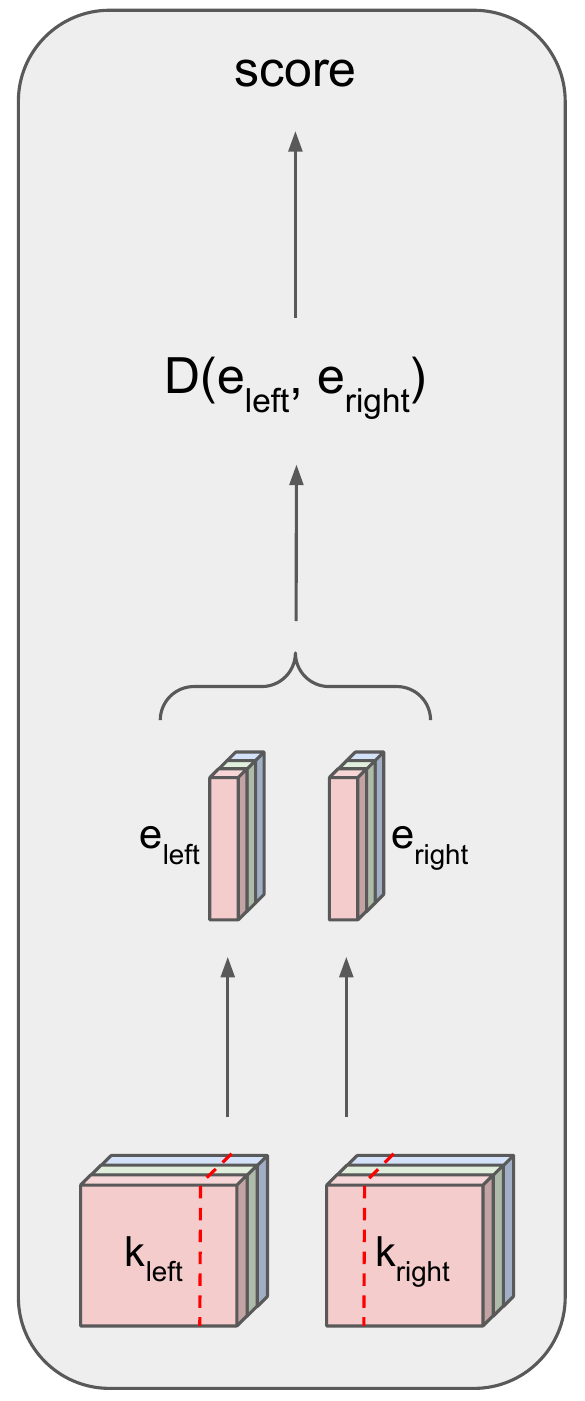}
    \caption{}
  \end{subfigure}
  \hfill
  \begin{subfigure}{0.3\linewidth}
  \centering
    \includegraphics[width=0.7\linewidth]{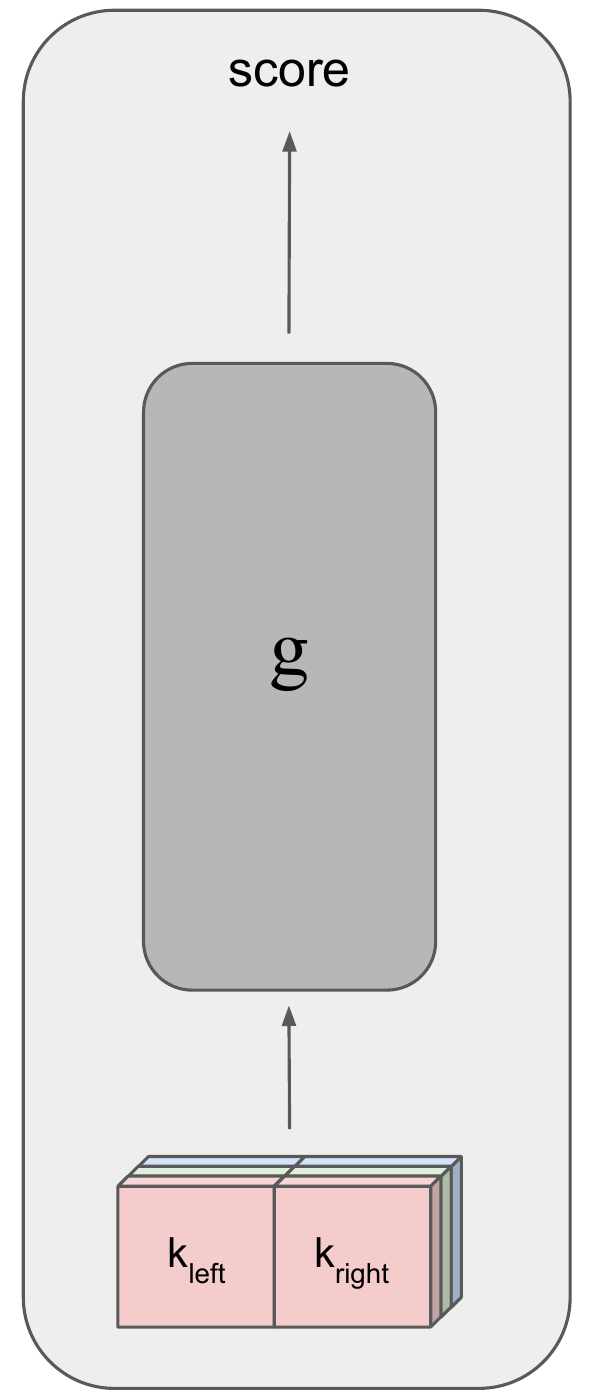}
    \caption{}
  \end{subfigure}
  \hfill
  \begin{subfigure}{0.321\linewidth}
  \centering
    \includegraphics[width=0.7\linewidth]{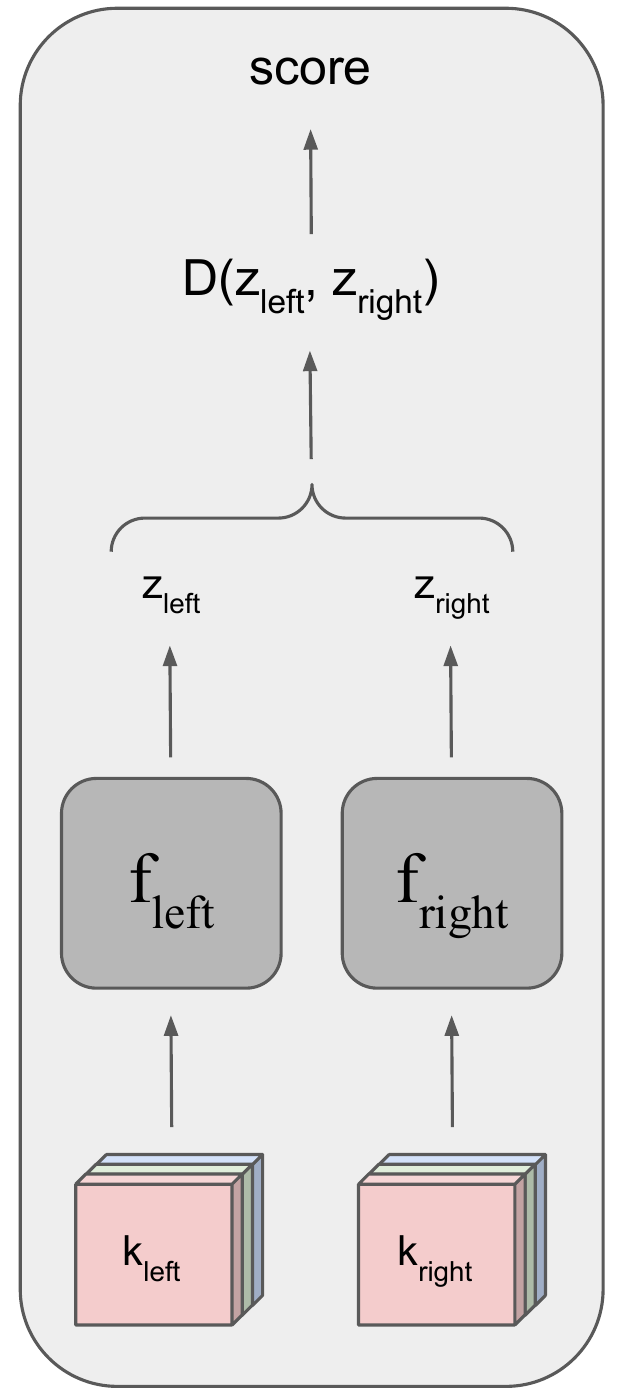}
    \caption{}
  \end{subfigure}
  \caption{Schematic architectures for CM computation: (a) Traditional, (b) NN-based, and (c) TEN, where $D$ represents a general distance measure function (not necessarily the same in (a) and (c)).}
  \label{fig:cm_comparison}
\end{figure}

Instead, we propose a hybrid approach, which uses an NN-based model for extracting the semantic information of a piece $k$ into a latent space $\mathcal{Z}$, and combining it with a relatively simple distance measure $D$, to determine how two latent vectors are compatible to each other.
The idea is that the NN-based model $f: \mathbb{R}^{P\times P\times C} \to \mathbb{R}^d$ extracts, in this manner, the most relevant features of the entire piece content with respect to each of its edges.
Obviously, we would need two networks to obtain the above representations for a given pair of pieces, {\ie} a left network $f_{left}$ and a right network $f_{right}$ for the left piece and right piece, respectively.
Similarly to the traditional CM computations of a given pair, which work on edge pixels, 
we can now apply simply a distance measure $D(z_l, z_r): \mathcal{Z} \times \mathcal{Z} \to \mathbb{R}$ with respect to the latent representations $z_l=f_{left}(k_l)$ and $z_r=f_{right}(k_r)$ of the pieces.
(Note, that physically $z_l$ corresponds to the right edge of $k_l$, and $z_r$ corresponds to the left edge of $k_r$.)
In our case, the dimensionality of the latent space is (expected to be) much smaller than that of the traditional approaches, as these methods operate on (all of) the raw edge pixels.
See Figure~\ref{fig:cm_comparison} for the different schematic architectures of the above main approaches.

\subsection{Architecture}
\label{sec:ten_architecture}
As indicated before, TEN consists of twin CNN-based models, sharing the same architecture but with different learned weights. Each twin network of our small TEN version consists of four convolutional layers and two interleaving max-pooling layers~\cite{nagi2011max}, using non-linear ReLU activation layers~\cite{nair2010rectified}, and a fully-connected layer on top projecting the CNN outputs onto a latent vector in $\mathcal{Z}$. See Table~\ref{tab:cnn_architecture} below for a detailed architecture. 

\newcolumntype{A}{>{\centering}p{4.5cm}}
\newcolumntype{B}{>{\centering}p{3cm}}
\newcolumntype{C}{>{\centering\arraybackslash}p{3cm}}
\begin{table}[h]
    \centering
    \caption{Our core architecture of both $f_{left}$ and $f_{right}$ embedding models; total number of model parameters depends on piece size and embedding dimension, {\ie} (28, 28, 3) and $d = 40$, in our case; training our NN-based E2E model was done with input size of $(28, 56, 3)$ and $d=1$.}
    \vspace{0.25cm}
    \begin{tabular}{|A|B|C|}
        \hline
        \multirow{2}{*}{\textbf{Layer Type}} & \multirow{2}{*}{\textbf{Output Dim}} & \multirow{2}{*}{\textbf{\# Params}} \\
        && \\
        \hline \hline
        Input & (H, W, 3) & - \\
        \hline
        Conv2D, filter=(3, 3) & (H, W, 64) & 1,728 \\
        \hline
        Conv2D, filter=(3, 3) & (H, W, 128) & 73,728 \\
        \hline
        MaxPool2D, filter=(2, 2) & (H/2, W/2, 128) & - \\
        \hline
        Conv2D, filter=(3, 3) & (H/2, W/2, 256) & 294,912 \\
        \hline
        MaxPool2D, filter=(2, 2) & (H/4, W/4, 256) & - \\
        \hline
        Conv2D, filter=(3, 3) & (H/4, W/4, 512) & 1,179,648 \\
        \hline
        Linear & ($d$, ) & $32\rm{H}\rm{W}d$ \\
        \hline \hline
        \multirow{2}{*}{\textbf{Total parameters:}} & \multicolumn{2}{c|}{\multirow{2}{*}{1,550,016 + $32\rm{H}\rm{W}d$}} \\
        & \multicolumn{2}{c|}{}\\
        \hline
        
    \end{tabular}
    \label{tab:cnn_architecture}
\end{table}

To extract even higher-quality embedding vectors, we also propose a larger version,  TEN-Large, in which each twin is an ensemble of four CNN-based networks with the same architecture as in Table~\ref{tab:cnn_architecture}, but with different inputs. Following~\cite{rika2019gecco}, each sub-network receives a unique color channel ({\ie} red, green or blue), and the fourth network receives the entire (3-channel) RGB input.
In principle, we observed that training an embedding model for each channel separately yields unique embedding features per channel, which results in added information to the ensemble architecture and in overall enhanced representation.
TEN-Large has 20.4M parameters, {\ie} 4 times as many as the the original TEN, as well as 4 times the number of embeddings. The pairwise compatibility using TEN-Large is given by
\begin{equation}
\label{eq:ten_large_inference}
\text{CM}_{\text{TEN-Large}}(k_l, k_r) = \frac{1}{4} \sum_{i} {D(f_{left}^{i}(k_l), f_{right}^{i}(k_r))},
\end{equation}
where $i$ represents a sub-network index of the ensemble.


\subsection{Training with Embeddings}
\label{sec:loss_function}

A procedure for computing a pairwise CM should return the lowest distance between an \textit{anchor} boundary and its true adjacent piece boundary ({\ie} \textit{positive pair}), relatively to the distances between the \textit{anchor} boundary and any piece edge in the puzzle ({\ie} \textit{negative pairs}).
In order to train our twin embedding networks to do so, we used  \textit{triplet-loss}~\cite{facenet2015cvpr} defined by the following objective function:
\begin{equation}
\label{eq:triplet_loss}
    \mathcal{L}_{triplet} = \max(0, D(z_a, z_p) - D(z_a, z_n) + \gamma),
\end{equation}
where $\gamma$ is set to $1$, $z_a$ denotes the embedding vector of $f_{left}$ (as the anchor edge is always located on the left-hand side), and where $z_p$ and $z_n$ denote the embedding vectors of $f_{right}$ (for the positive and negative edges, respectively), as they are located on the right-hand side.
Plausible distance measures could be the inverse cosine similarity, $L_1$ distance, Euclidean distance, or any other distance measure between two vectors.
In principle, we found that picking triplet piece edges from the same puzzle leads to better performance. This could be attributed to the notion that negative piece edges from the same puzzle are more likely to be similar to their positive counterparts, thereby forcing the twin embedding networks to extract more discriminative embedding vectors.


\subsection{Post Processing}
\label{sec:post_processing}

Note that TEN will return different dissimilarity scores for a given pair and its rotated version by $180^{\circ}$, since these pairs do not look the same from the network's perspective. Of course, obtaining an asymmetric compatibility measure for the very same pair does not make sense.  
To remedy this undesired effect, we apply post processing after computing all pairwise permutations $(k_i, k_j)$.
\begin{equation}
\label{eq:minmax_scaling}
    CM'(k_i,k_j) = \dfrac{CM(k_i,k_j) - min(CM(k_i,*))}{max(CM(k_i,*)) - min(CM(k_i,*))}
\end{equation}
We then ensure symmetry for the same pair of pieces by defining:
\begin{equation}
    CM''(k_i,k_j) = CM''(k_j,k_i) = \dfrac{CM'(k_i,k_j) + CM'(k_j,k_i)}{2}
\end{equation}


\subsection{Type-1 and Type-2 Puzzles}
It is common to consider two levels of difficulty for the JPP, namely \textit{Type-1} and \textit{Type-2}.
Type-1 assumes that only piece locations are unknown, while their orientations are known and fixed. Type-2 relaxes this assumption, {\ie} piece orientations, as well piece locations, are unknown.
Note that a CM for Type-1 needs to discriminate, for each anchor piece $k_a$, between its positive piece $k_p$ ({\ie} ground-truth neighbor) and all of the negative candidates $k_n$, from a total number of $N-1$ candidates. For Type-2, the total number of candidates rises to $4(N-1)$, as there are 4 possible orientations for each piece.
From a combinatorial point of view, there are essentially $4N$ anchor edges for an $N$-piece puzzle, such that each anchor edge has $N-1$ and $4(N-1)$ possible matching edges, for Type-1 and Type-2, respectively. In other words, a CM needs to be computed $4N(N-1)$ and $16N(N-1)$ times, {\ie} roughly $4N^2$ and $16N^2$ times for Type-1 and Type-2, respectively.


\subsection{Fast Inference}
\label{sec:fast_inference}
Let us now examine theoretically the significant reduction in time complexity due to the use of embeddings, as described in Section~\ref{sec:twin_embedding_nets}.
Assume a puzzle of $N$ pieces with unknown piece position and orientation. As previously indicated, there are $16N^2$ possible combinations for this Type-2 variant, {\ie} a CM module must be executed for all of these $16N^2$ combinations.
Computing CMs for this large number of combinations by the classical methods may not have been a major issue, due to their relatively simple computation.
In contrast, NN-based CMs are very computationally intensive, which makes them rather slow and virtually impractical for puzzles containing more than a few hundred pieces.
Fortunately, our proposed TEN framework offers a highly-efficient alternative. Observe that each of the $f_{left}$ and $f_{right}$ networks needs to execute only $4N$ times to capture all of the 4-edge representations of each puzzle piece.

After TEN is executed $4N$ times, we save all of these embedding representations in two tensors of size ($N$, $4$, $d$), $T_{left}$ and $T_{right}$, where $d$ is the embedding dimensionality. At this point, we merely employ a very simple distance measure on the embeddings to obtain the CMs of all pairwise puzzle pieces. Although the distance measure needs to be computed $16N^2$ times, the running time it takes is by far smaller than employing an NN-based CM the same number of times, which involves millions/billions of FLOPS (for each pair).
In fact, the larger the puzzle is, the larger the speedup gain obtained over an NN-based CM, as will be demonstrated in Subsection~\ref{sec:inference_times}.


\section{Experimental Results}
\label{sec:experimental_results}
During our extensive experiments, we trained and evaluated the following three models: (1) Regular version of our TEN architecture, (2) ensemble version, TEN-Large, which contains four times as many parameters as TEN, and (3) pairwise, NN-based end-to-end (E2E) CM, as depicted in Figure~\ref{fig:cm_comparison}(b).
For the latter model, we used basically the same architecture presented schematically in Table~\ref{tab:cnn_architecture}, but instead of $d$ outputs,
this network outputs a single value, which indicates the degree of compatibility of a given pair of pieces.

We have experimented intentionally with an NN-based model, as well, to study not only the actual speedup gain due to TEN during inference, but to assess the expected accuracy gap as a performance reference for future research.

We trained all of the above networks with the same recipe, including an Adam optimizer~\cite{kingma2017adam}, a learning rate of 1E-4 (with a decay factor of $0.9$, in case the loss did not decrease for 5 epochs), a batch size of 64, and an epoch of $5000$ iterations.
To obtain the final results for TEN, TEN-Large, and NN-based E2E, we trained them over $600$ epochs.
TEN and TEN-Large were trained with \textit{triplet-loss}, while the NN-based CM was trained with the \textit{binary cross entropy loss} (labeling positive pairs as ``0'' and negative pairs as ``1'').


\subsection{Datasets}
\label{sec:datasets}

We used the 800 training images of the DIV2K~\cite{Agustsson_2017_CVPR_Workshops} dataset to train our model.
For the evaluation part, we used the 100 validation images of DIV2K, 100 images of the PIRM dataset~\cite{pirm2018cvpr}, and 20 images of the MIT dataset (see Cho~{\etal}~\cite{conf/cvpr/ChoAF10}.
We selected the above three different datasets, which are publicly available and commonly used in computer vision, to experiment with a decent, diverse collection of puzzles to establish our method's reliability.
To generate a degraded version of the original datasets, we then cropped each puzzle image into pieces of size $28 \times 28$ pixels, and deleted the boundary pixels of each piece, yielding an erosion effect of 13.8\% (with respect to the piece size).

As indicated before, we experimented with eroded-boundary pieces, as this problem instance serves as a testbed for various practical, real-world JPPs, which must deal with damaged tiles around the edges. We generated only 1-pixel erosions, as this minimal degradation already resulted in poor CMs. (See Table~\ref{tab:top_i_acc} for specific results.)
To evaluate the classical CMs on 1-pixel erosions, in a meaningful manner, we disposed the entire piece frame and considered only its remaining content ({\ie}, instead of the original $28 \times 28$-pixel pieces, we considered their corresponding eroded pieces of size $26 \times 26$ pixels).
On the other hand, we retained the original piece size for all trainable CMs ({\ie} TEN, TEN-Large, and NN-based E2E), such that each piece includes its eroded frame.
Also, for the classical SSD, PBC, and $L_1$ CMs we retained the image representation in LAB color space (according to the specification of these methods), while using an RGB representation for all other CMs.


\subsection{Dimensionality Effect}
\label{sec:dimensionality_effect}
In general, an embedding dimensionality is one of its main characteristics. To determine practically the desired embedding dimensionality for an adequate representation of a piece edge $e$, we trained TEN on various dimensions, {\ie} $d \in \{ 10, 20, 40, 80 \}$.
We stopped at $d=80$, since TEN did not seem to yield better \textit{Top-1 accuracy} on our test set, using a larger number of dimensions. (See exact definition of Top-1 accuracy in Subsection~\ref{sec:top1_eval}.) 
Indeed, empirical results consistently showed that using $d=40$ was satisfactory, regardless of the distance measure used (see Table~\ref{tab:dimension_comparison}).

\newcolumntype{A}{>{\centering}p{3cm}}
\newcolumntype{B}{>{\centering}p{1.5cm}}
\newcolumntype{C}{>{\centering\arraybackslash}p{1.5cm}}
\begin{table}
\centering
\caption{Dimensionality effect: $d > 40$ does not seem to gain higher Top-1 accuracy for both distance measures used, running TEN
on half epochs (rather than on full training phase).}
\vspace{0.25cm}
\begin{tabular}{ |A|B|B|B|C| }
\hline
\multirow{2}{*}{\textbf{Distance Measure}} & \multicolumn{4}{c|}{\textbf{Embedding Size - $d$}} \\
\cline{2-5}
& \textbf{10} & \textbf{20} & \textbf{40} & \textbf{80} \\ 
\hline \hline
\small{$1 - \rm{cosine}$} & \small{31.1\%} & \small{37.9\%} & \small{40.7\%} & \small{40.5\%}
\\
\hline
\small{Euclidean distance} & \small{46.3\%} & \small{51.4\%} & \small{54.4\%} & \small{54.7\%}
\\
\hline
\end{tabular}
\label{tab:dimension_comparison} 
\end{table}


\subsection{Distance Measure}
\label{sec:distance_metric}
Recall, computing the compatibility score between two embeddings $z_l$, $z_r$ requires the use of a distance function $D: \mathcal{Z} \times \mathcal{Z} \to \mathbb{R}$. Technically, $D$ can be any desired function, including an NN-based one; however, to accelerate the computation as much as possible, we have focused on fast and simple distance measures.

\begin{itemize}
    \item[$\bullet$] \textbf{Cosine similarity:} This embedding similarity is very common these days, in particular among the \textit{self-supervised learning} field.
    Let $z_l$ and $z_r$ denote the embeddings obtained. Then the cosine similarity is given by
    \begin{equation}
        cosine(z_l, z_r) = \frac{z_l \cdot z_r}{\| z_l \|  \| z_r \|}
    \end{equation}
    In its raw form, the cosine projects all distances to the range $[-1, 1]$, where the closest embeddings are assigned a score of $1$ and vice versa. To force the cosine similarity to assign a near-zero value to similar embeddings, we define
    \begin{equation}
        D(z_l, z_r) = 1 - cosine(z_l, z_r),
    \end{equation}
    such that all distances now lie in the range
    $[0, 2]$.
    \item[$\bullet$]\textbf{$\text{L}_p$ distance:} This is a family of distance measures which differ from one another by the parameter $p$. In this work we examined three $p$ values  $\in \{ 1,2,3 \}$, which correspond to the well-known $L_1$, $L_2$ (Euclidean distance), and $L_3$ metrics.
    In general, the $L_p$ distance between two embeddings is calculated according to
    \begin{equation}
        L_p(z_l, z_r) = \left( \sum_{i=1}^{d}{|z_{l}[i] - z_{r}[i]|^p} \right)^{1/p}
    \end{equation}
\end{itemize}

In all of the experiments, we trained our models with the above distance measures for $d = 40$. A comparative performance is given in Table~\ref{tab:distance_metrics_comparison} with respect to Top-1 accuracy (see definition below). The results indicate that the $L_p$ norm is preferable to the cosine similarity. More specifically, the best results were achieved using the $L_2$ Euclidean distance.

\newcolumntype{A}{>{\centering}p{4cm}}
\newcolumntype{C}{>{\centering\arraybackslash}p{4cm}}
\begin{table}
\centering
\caption{Top-1 accuracies of our model (trained on half epochs) for four simple, fast distance measures and embedding size of 40; $L_2$ metric achieved highest score.}
\vspace{0.25cm}
\begin{tabular}{|A|C|}
\hline
\multirow{2}{*}{\textbf{Distance Measure}} & \multirow{2}{*}{\textbf{Top-1 Accuracy}} \\ 
&  \\ 
\hline \hline 
$1 - \rm{cosine}$ & 40.7\% \\ \hline
$L_1$ & 51.8\% \\ \hline
$L_2$ (Euclidean distance) & 54.4\% \\ \hline
$L_3$ & 54\% \\ \hline
\end{tabular}
\label{tab:distance_metrics_comparison} 
\end{table}


\subsection{Performance Results}
\label{sec:performance_results}


\subsubsection{Top-1 Accuracy}
\label{sec:top1_eval}
Ideally, given an anchor piece edge, we would like an algorithm that assigns the best pairwise CM to the true adjacent edge among all other piece edges in the puzzle. The most common criterion to quantitatively assess the performance of any such CM procedure is the Top-1 classification rate.

We define a pair $(z_a, z_p)$ of an (embedded) anchor edge and its (embedded) true adjacent edge to be Top-$i$ compatible (with respect to a given CM procedure) if its computed CM is greater than the $i$-th computed CM between the (embedded) anchor edge and any (embedded) negative edge $(z_a, z_n)$, $\forall z_n \neq z_a, z_p$.
Ideally, Top-1 should approach 100\% accuracy for a perfect CM module. Practically, the higher it is, the more accurate the CM computation.

We computed Top-1 accuracies on the entire test set of 100 puzzles, each containing 800-1000 pieces.
As the results indicate, NN-based E2E performs best, for now, with Top-1 accuracies of 71.4\% and 64.1\% for type-1 and type-2, respectively.
Our TEN-Large yields, though, second-best CM scores with Top-1 accuracies of 64.6\% and 56.8\% for Type-1 and Type-2, respectively.
This gap might be attributed to the intuitive insight that an NN-based model ``sees'' an entire pair before assessing the extent of compatibility between the two pieces. In contrast, the twin models $f_{left}$ and $f_{right}$ try to predict, essentially, each other's embedding, which is a harder task. 

\begin{table}
\centering
\setlength{\tabcolsep}{7pt}
\caption{Top-1 accuracies for various CMs: Our embedding-based CM is more accurate than all classical methods tested (best results appear in bold, and second best are underlined); for now, NN-based E2E yields higher rates.}
\vspace{0.25cm}
\begin{tabular}{ l c c c c c c c }
\toprule

\multirow{2}{*}{\textbf{Method}} & \multicolumn{3}{c}{\textbf{Type-1}} & & \multicolumn{3}{c}{\textbf{Type-2}} \\
\cmidrule(l){2-4} \cmidrule(l){6-8}

& \textbf{DIV2K} & \textbf{PIRM} & \textbf{MIT} & & \textbf{DIV2K} & \textbf{PIRM} & \textbf{MIT} \\

\midrule

SSD & 37.4\% & 41\% & 42\% && 29\% & 32.2\% & 33.5\% \\
PBC & 46.4\% & 49.1\% & 43.7\% && 37.5\% & 39.3\% & 34.1\% \\
$L_1$ & 47\% & 49.8\% & 46.6\% && 38.7\% & 40.7\% & 37.6\% \\
MGC & 56.3\% & 59.5\% & 53.5\% && 45.4\% & 47.5\% & 41.1\% \\
TEN (ours) & \underline{59.4\%} & \underline{61\%} & \underline{55.2\%} && \underline{50.5\%} & \underline{52.5\%} & \underline{46.6\%} \\
TEN-L (ours) & \textbf{64.6\%} & \textbf{65.4\%} & \textbf{59.1\%} && \textbf{56.8\%} & \textbf{57.5\%} & \textbf{51.5\%} \\

\midrule

NN-based E2E & 71.4\% & 72.6\% & 69.4\% && 64.1\% & 64.2\% & 60.9\% \\

\bottomrule
\end{tabular}
\label{tab:top_i_acc}
\end{table}


\begin{figure}[]
\centering
    \begin{tabular}{c c c}
    
        \includegraphics[width=0.288\linewidth]{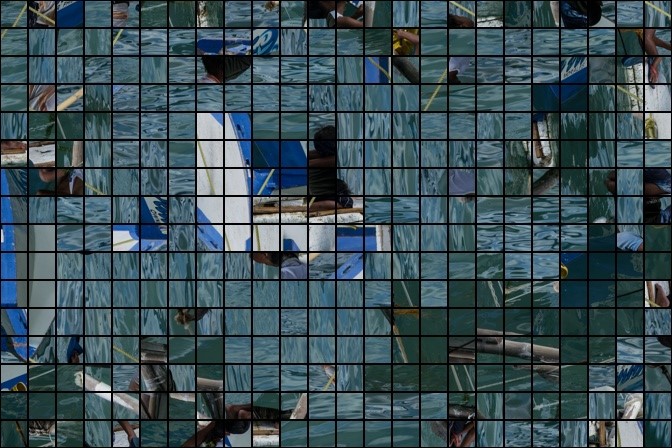} &
        \includegraphics[width=0.288\linewidth]{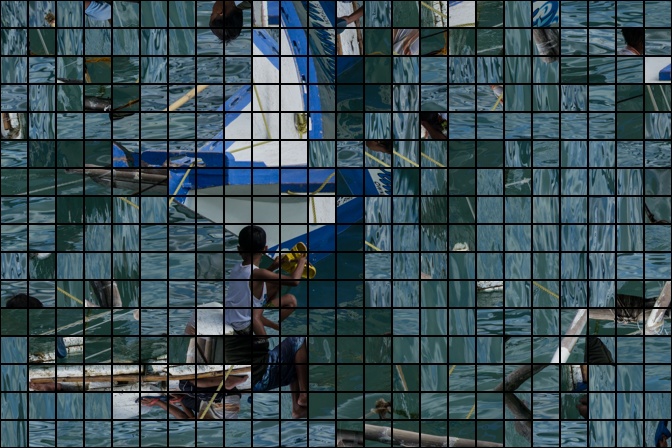} &
        \includegraphics[width=0.288\linewidth]{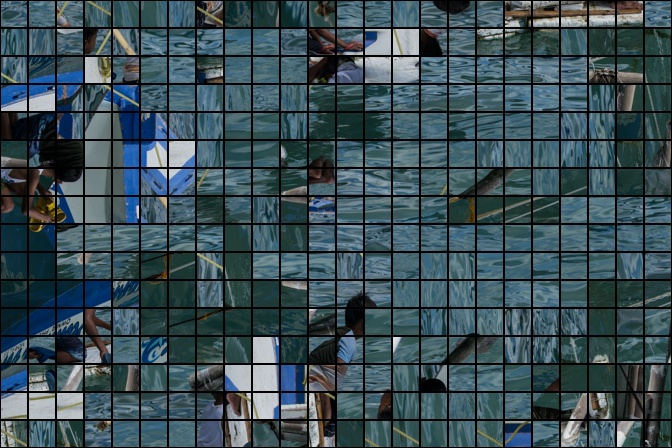}
        \\ \footnotesize{SSD 16.6\%} & \footnotesize{PBC 30.8\%} & \footnotesize{$L_1$ 25.7\%} \\
        \includegraphics[width=0.288\linewidth]{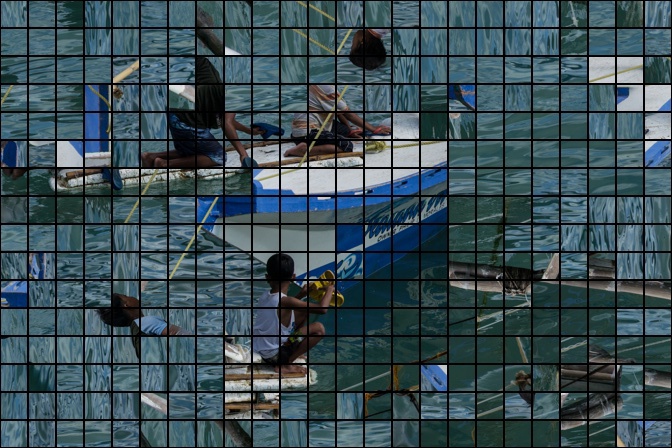} &
        \includegraphics[width=0.288\linewidth]{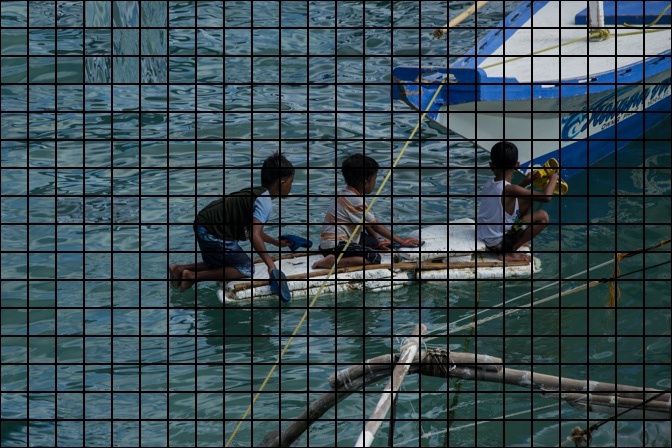} &
        \includegraphics[width=0.288\linewidth]{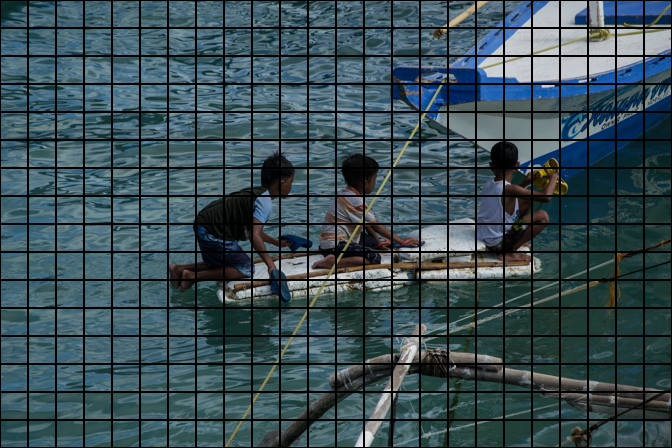}
        \\ \footnotesize{MGC 34.1\%} & \footnotesize{TEN-Large 80.4\% (ours)} & \footnotesize{Ground Truth} \\
        
        \toprule \\
        
        \includegraphics[width=0.288\linewidth]{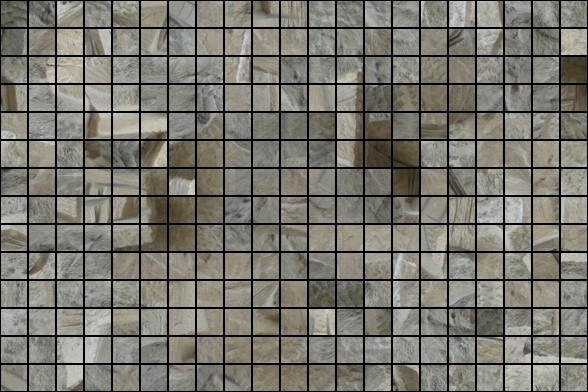} &
        \includegraphics[width=0.288\linewidth]{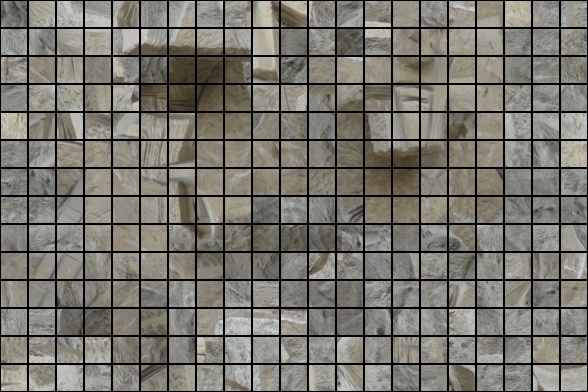} &
        \includegraphics[width=0.288\linewidth]{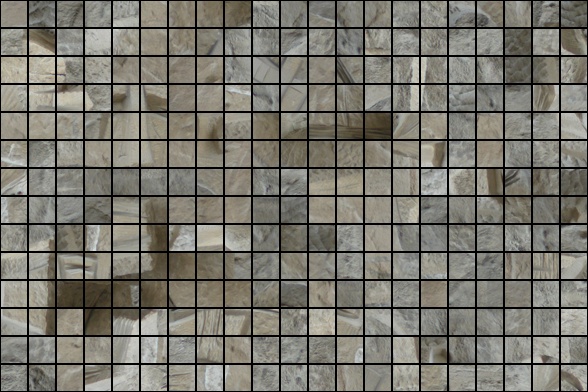}
        \\ \footnotesize{SSD 10.1\%} & \footnotesize{PBC 13.2\%} & \footnotesize{$L_1$ 12.1\%} \\
        \includegraphics[width=0.288\linewidth]{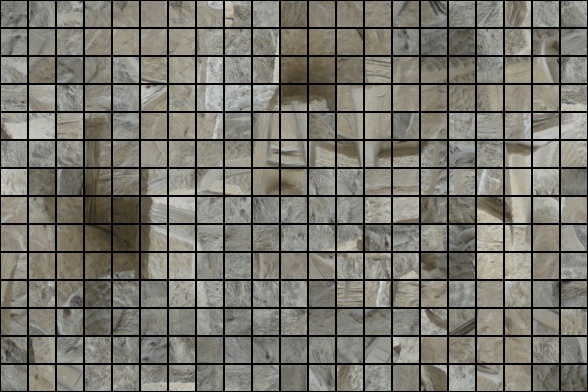} &
        \includegraphics[width=0.288\linewidth]{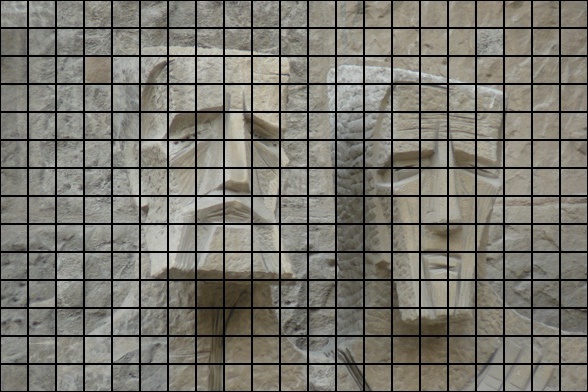} &
        \includegraphics[width=0.288\linewidth]{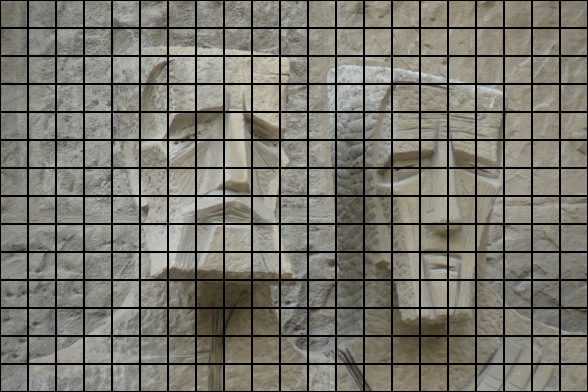}
        \\ \footnotesize{MGC 18.3\%} & \footnotesize{TEN-Large 82.3\% (ours)} & \footnotesize{Ground Truth} \\
        
        \toprule \\
        
        \includegraphics[width=0.288\linewidth]{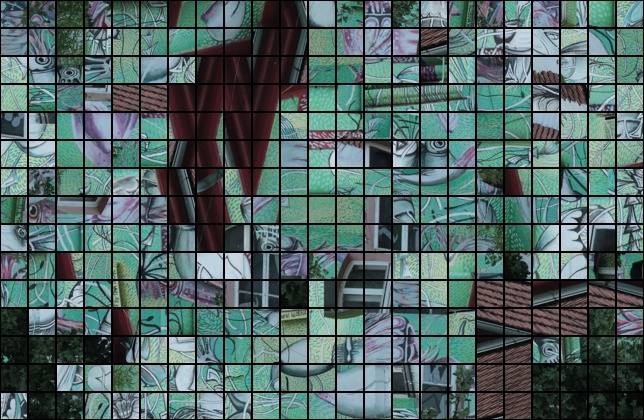} &
        \includegraphics[width=0.288\linewidth]{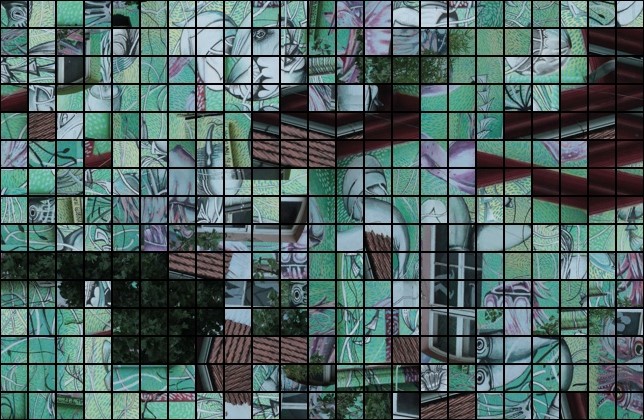} &
        \includegraphics[width=0.288\linewidth]{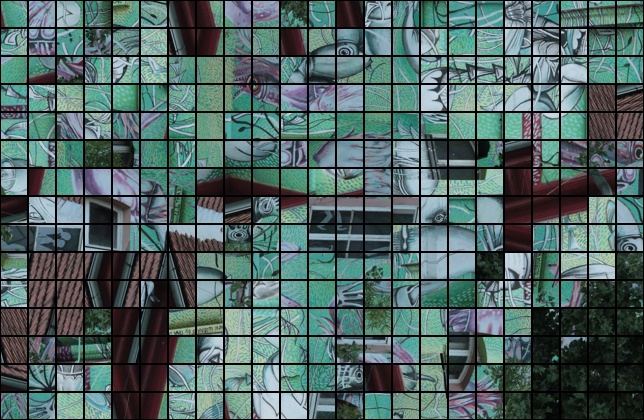}
        \\ \footnotesize{SSD 15\%} & \footnotesize{PBC 23.8\%} & \footnotesize{$L_1$ 18.7\%} \\
        \includegraphics[width=0.288\linewidth]{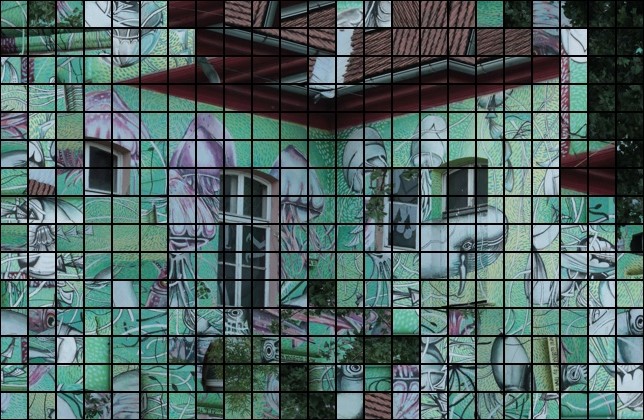} &
        \includegraphics[width=0.288\linewidth]{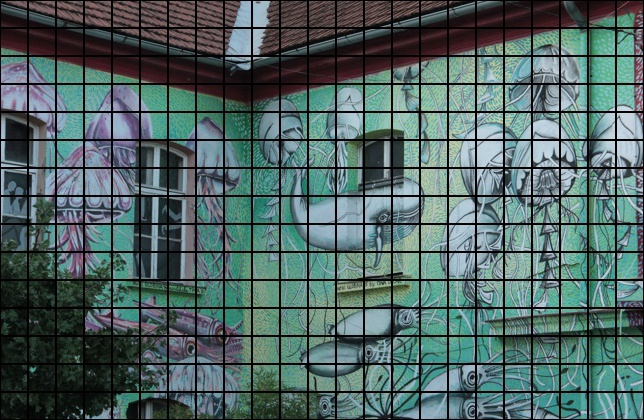} &
        \includegraphics[width=0.288\linewidth]{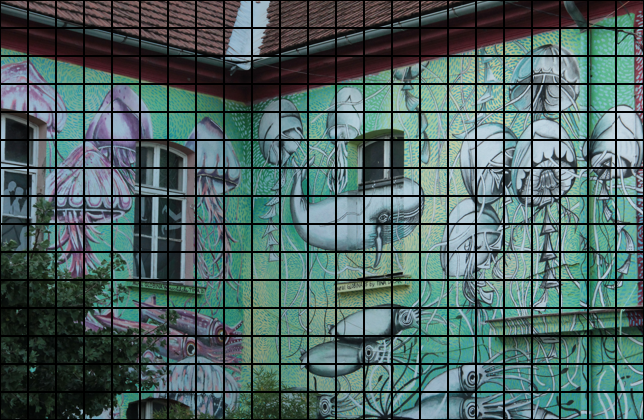}
        \\ \footnotesize{MGC 39.6\%} & \footnotesize{TEN-Large 93.7\% (ours)} & \footnotesize{Ground Truth} \\
    
    \end{tabular}
\caption{Sample of reconstructions due to~\cite{rika2019gecco} for Type-2 variant; our TEN-Large model exhibits superior performance to that of classical methods.}
\label{fig:reconstruction_comparison}
\end{figure}

\subsubsection{Black-Box Reconstruction}
\label{sec:reconstruction_eval}
Another way to conduct a comparative performance evaluation is to supplement each CM module with the same reconstruction algorithm,
and test which CM, coupled with this reconstruction ``black box'', contributes the most to the reassembled puzzle.
We chose two reconstruction algorithms: (1)  Gallagher's greedy reconstruction algorithm, whose code is publicly available and which is commonly used in many other studies, and (2) the genetic algorithm (GA)-based reconstruction scheme due to~\cite{rika2019gecco}. Unlike Gallagher's solver, the latter scheme searches for a global optimization by correcting piece misplacements at early stages of the reconstruction.
To assess the reconstruction accuracy, we apply the \textit{neighbor accuracy} criterion. (For the GA-based solver we actually report the average neighbor accuracy over 10 runs due to its non-deterministic behavior.)
We experimented with all of the traditional CMs, as well as our proposed TEN and TEN-Large, and recorded for each method the average neighbor accuracy (over the test images of each dataset), after employing the above reconstruction solvers.
Note that the reconstruction algorithms were employed uniformly, just for an additional comparative evaluation, {\ie} the main point here is not necessarily the absolute neighbor accuracy obtained, but rather the significant improvement due to the novel CM model introduced, relatively to the various traditional measures tested.

The neighbor accuracy results (for the evaluated CMs coupled with the reconstruction algorithms) are summarized in Table~\ref{tab:neighbor_acc} for both Type-1 and Type-2 variants.
In general, the results reveal the higher reconstruction accuracy due to the GA-based solver.
More importantly, using TEN-Large leads to improved accuracy of up to $8.5\%$ and $16.8\%$, respectively, for Type-1 and Type-2, relatively to the best known classical MGC.

A visual reconstruction comparison between the classical CMs and our best proposed model, TEN-Large, is shown in Figure~\ref{fig:reconstruction_comparison} on a few selected images.

\begin{table}
\centering
\setlength{\tabcolsep}{6.5pt}
\caption{Reconstruction neighbor accuracy due to evaluated CMs with Gallagher's greedy solver~\cite{gallagher2012jigsaw} and GA-based solver~\cite{rika2019gecco}; average accuracy (over 10 runs for each puzzle) reported for latter reconstruction; TEN (and TEN-Large) models are superior to classical methods for both Type-1 and Type-2 (best results appear in bold, and second best are underlined).}
\vspace{0.25cm}
\begin{tabular}{ l c c c c c c c }

\multicolumn{8}{c}{\textbf{Type-1}} \\

\toprule

\multirow{2}{*}{\textbf{Method}} & \multicolumn{3}{c}{\textbf{Gallagher's solver}} & & \multicolumn{3}{c}{\textbf{GA-based solver}} \\
\cmidrule(l){2-4} \cmidrule(l){6-8}

& \textbf{DIV2K} & \textbf{PIRM} & \textbf{MIT} & & \textbf{DIV2K} & \textbf{PIRM} & \textbf{MIT} \\

\midrule

SSD & 28.7\% & 32.4\% & 34.5\% && 27.9\% & 34\% & 32.4\% \\
PBC & 43.7\% & 47.1\% & 36.4\% && 49.3\% & 53.2\% & 44.7\% \\
$L_1$ & 42.2\% & 46\% & 41.3\% && 43.6\% & 50.9\% & 43.2\% \\
MGC & \underline{54.7\%} & \textbf{60.2\%} & \underline{49.4\%} && 64.4\% & 71.4\% & 58.2\% \\
TEN (ours) & 52.9\% & 53.1\% & 44.9\% && \underline{67.5\%} & \underline{71.5\%} & \underline{61.2\%} \\
TEN-Large (ours) & \textbf{60.7\%} & \underline{59.8\%} & \textbf{53.4\%} && \textbf{72.9\%} & \textbf{76.8\%} & \textbf{65.1\%} \\

\midrule

NN-based E2E & 75.4\% & 76.1\% & 69\% && 82.5\% & 90.8\% & 87.3\% \\

\bottomrule

\\

\multicolumn{8}{c}{\textbf{Type-2}} \\

\toprule

\multirow{2}{*}{\textbf{Method}} & \multicolumn{3}{c}{\textbf{Gallagher's solver}} & & \multicolumn{3}{c}{\textbf{GA-based solver}} \\
\cmidrule(l){2-4} \cmidrule(l){6-8}

& \textbf{DIV2K} & \textbf{PIRM} & \textbf{MIT} & & \textbf{DIV2K} & \textbf{PIRM} & \textbf{MIT} \\

\midrule

SSD & 7.3\% & 9.8\% & 10.3\% && 16.3\% & 18.4\% & 18.3\% \\
PBC & 13.8\% & 16.5\% & 9.5\% && 26.1\% & 30.5\% & 20.9\% \\
$L_1$ & 12.2\% & 15\% & 10.7\% && 25.6\% & 28.2\% & 22\% \\
MGC & 19.7\% & 22.7\% & 12.8\% && 40.2\% & 47.5\% & 32.3\% \\
TEN (ours) & \underline{21.8\%} & \underline{23.4\%} & \underline{16.6\%} && \underline{46.9\%} & \underline{50.9\%} & \underline{39.6\%} \\
TEN-Large (ours) & \textbf{30.2\%} & \textbf{33.6\%} & \textbf{22.7\%} && \textbf{55.5\%} & \textbf{59.4\%} & \textbf{49.1\%} \\

\midrule

NN-based E2E & 48.5\% & 49.9\% & 39\% && 73.3\% & 79.7\% & 74.5\% \\

\bottomrule
\end{tabular}
\label{tab:neighbor_acc}
\end{table}

\newcolumntype{A}{>{\centering}p{2cm}}
\newcolumntype{B}{>{\centering}p{2cm}}
\newcolumntype{C}{>{\centering\arraybackslash}p{2cm}}
\begin{table}
    \centering
    \caption{Wall-clock inference times (\textbf{in seconds}) of our proposed embedding-based CMs compared to NN-based E2E; TEN (and TEN-Large) boost inference speed by orders of magnitude as more pieces are added to puzzle.}
    \vspace{0.25cm}
    \begin{tabular}{ A B B C }
        \textbf{\# Pieces} & \textbf{TEN} & \textbf{TEN-Large} & \textbf{E2E} \\
        \midrule
        \textbf{100} & $7.0 \times 10^{-2}$ & $1.8 \times 10^{-1}$ & $3.2 \times 10^{1}$ \\
        \textbf{200} & $1.4 \times 10^{-1}$ & $3.9 \times 10^{-1}$ & $1.2 \times 10^{2}$ \\
        \textbf{400} & $3.6 \times 10^{-1}$ & $8.9 \times 10^{-1}$ & $4.9 \times 10^{2}$ \\
        \textbf{800} & $1.0 \times 10^{0}$ & $2.2 \times 10^{0}$ & $1.9 \times 10^{3}$ \\
        \textbf{1600} & $3.2 \times 10^{0}$ & $6.2 \times 10^{0}$ & $7.9 \times 10^{3}$ \\
        \textbf{3200} & $1.1 \times 10^{1}$ & $1.9 \times 10^{1}$ & $3.1 \times 10^{4}$ \\
        \bottomrule
    \end{tabular}
    \label{tab:inference_times}
\end{table}


\subsubsection{Inference Times}
\label{sec:inference_times}

Table~\ref{tab:inference_times} gives the actual running times for computing all of the $16N^2$ combinations of a single puzzle with $N$ pieces, on average.
We experimented with 
$N \in \{ 100, 200, 400, 800, 1600, 3200 \}$, {\ie} with a growing number of pieces, to underscore the impact of our proposed architecture compared to the NN-based (E2E) method.
The running times due to the embedding (of TEN and TEN-Large) include the time to compute the final distance measures for all combinations.

\begin{figure}
    \centering
    \includegraphics[width=1.0\linewidth]{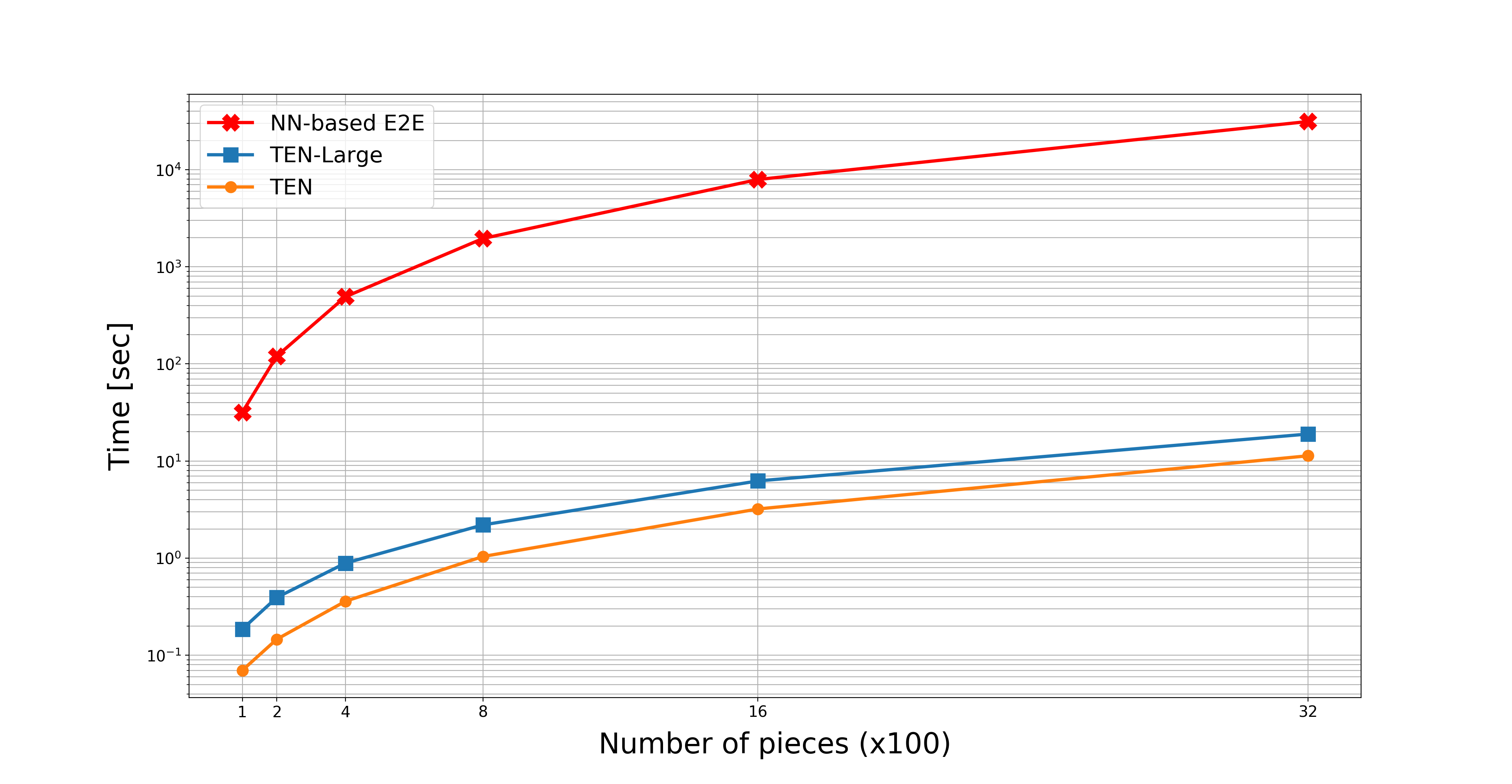}
    \caption{Comparative inference times of our new embedding-based CMs, TEN and TEN-Large, versus NN-based (E2E); computation time is dramatically reduced for $16N^2$ pairwise piece compatibilities.}
    \label{fig:inference_time}
\end{figure}

Note the speedup gain as a function of puzzle size.
Specifically, TEN/TEN-Large run 454/176 times faster than the NN-based model employed on 100-piece puzzles.
Moreover, our TEN/TEN-Large models become 2771/1657 times faster than E2E on puzzles with 3200 pieces, which makes TEN the fastest known NN-based CM module.
Figure~\ref{fig:inference_time} visualizes how fast our TEN and TEN-Large models compare to the NN-based (E2E) scheme.

Also, note that GPU/CPU utilization along with other factors like code optimization can cause changes in inference times as reported in Table~\ref{tab:inference_times}.
The reported inference times were measured on a modern PC with 3.5GHz CPU, 32GB RAM, and a single GPU with 11GB memory.

\section{Conclusions}
\label{sec:conclusions}
In this paper we presented a novel Twin Embedding (Neural) Network (TEN) framework, for computing very efficiently, yet fairly accurately, CMs for real-world JPP applications. The proposed embedding achieves a significant improvement in Top-1 accuracy compared to the classical CMs evaluated.
This innovation yields also an enhanced reconstruction of puzzles with eroded-boundary tiles, as was demonstrated by the comparison between different CMs coupled with the same black-box reconstruction. We have focused deliberately on a problem domain with such degraded puzzles, unlike most JPP works, which deal typically with perfect, synthetic puzzles, since it offers a challenging testbed for real-world scenarios, where puzzle pieces undergo severe degradation along their edges.

We believe that our novel embedding paradigm could also affect significantly the accelerated research and development of further JPPs containing (tens of) thousands of eroded pieces for highly challenging problems, without compromising much on inference speed, as opposed to computationally-intensive NN-based CM models which run slower by a few orders of magnitude.

Although for now NN-based E2E models are relatively more accurate, we intend to address this limitation, as part of future work, to further close the accuracy gap while maintaining fast running times.
For example, given the inherent redundancy between the $f_{left}$ and $f_{right}$ networks, in view of the same tasks performed by both models (albeit with respect to different edges), we intend to draw on this insight to derive a unified model for extracting bidirectional embeddings for a given piece.


\bibliographystyle{unsrt}  
\bibliography{egbib}

\end{document}